\documentclass[conference]{IEEEtran}
\IEEEoverridecommandlockouts
\usepackage{cite}
\usepackage{amsmath,amssymb,amsfonts}
\usepackage{algorithmic}
\usepackage{graphicx}
\usepackage{textcomp}
\usepackage{xcolor}
\def\BibTeX{{\rm B\kern-.05em{\sc i\kern-.025em b}\kern-.08em
    T\kern-.1667em\lower.7ex\hbox{E}\kern-.125emX}}
\usepackage{algorithm}
\usepackage{algorithmic}
\usepackage{amsmath}
\usepackage{graphicx}
\usepackage{subfig}
\usepackage{colortbl}
\usepackage[table]{xcolor}
\usepackage{verbatim}

\begin{document}

\title{AIVD: Adaptive Edge-Cloud Collaboration for Accurate and Efficient Industrial Visual Detection}

\author{
\IEEEauthorblockN{
Yunqing Hu\textsuperscript{1,3},
Zheming Yang\textsuperscript{1},
Chang Zhao\textsuperscript{1,3},
Qi Guo\textsuperscript{1,3},
Meng Gao\textsuperscript{2},
Pengcheng Li\textsuperscript{1,3},
Wen Ji\textsuperscript{1,2}
}
\IEEEauthorblockA{
\textsuperscript{1}Institute of Computing Technology, Chinese Academy of Sciences, Beijing, China\\
\textsuperscript{2}Institute of AI for Industries, Chinese Academy of Sciences, Nanjing, China\\
\textsuperscript{3}University of Chinese Academy of Sciences, Beijing, China
}
}

\maketitle

\begin{abstract}
Multimodal large language models (MLLMs) demonstrate exceptional capabilities in semantic understanding and visual reasoning, yet they still face challenges in precise object localization and resource-constrained edge-cloud deployment. To address this, this paper proposes the AIVD framework, which achieves unified precise localization and high-quality semantic generation through the collaboration between lightweight edge detectors and cloud-based MLLMs. To enhance the cloud MLLM's robustness against edge cropped-box noise and scenario variations, we design an efficient fine-tuning strategy with visual-semantic collaborative augmentation, significantly improving classification accuracy and semantic consistency. Furthermore, to maintain high throughput and low latency across heterogeneous edge devices and dynamic network conditions, we propose a heterogeneous resource-aware dynamic scheduling algorithm. Experimental results demonstrate that AIVD substantially reduces resource consumption while improving MLLM classification performance and semantic generation quality. The proposed scheduling strategy also achieves higher throughput and lower latency across diverse scenarios. 
\end{abstract}

\begin{IEEEkeywords}
Edge-cloud collaboration, multimodal large language models, industrial visual detection
\end{IEEEkeywords}

\section{Introduction}
\label{sec:intro}

In industrial visual detection scenarios, defects usually exhibit small scales, similar morphologies, and heavy background noise. These characteristics place stringent requirements on detection systems for precise localization and stable semantic classification. In practical production environments, lightweight detectors such as the YOLO series~\cite{1} and RT-DETR~\cite{2} are widely adopted due to their fast inference speed and ease of deployment. However, these models show clear limitations in semantic understanding, fine-grained classification, and structured description generation \cite{DBLP18}. When defect variations are extremely subtle, lightweight detectors often fail to distinguish them reliably using only local texture or shape cues. This limitation leads to frequent semantic confusion among visually similar defect categories. Moreover, most lightweight models can only output categorical labels. They cannot generate structured and interpretable explanations that are critical in industrial applications.

\begin{figure}[th]
    \centering
    \includegraphics[width=1\linewidth]{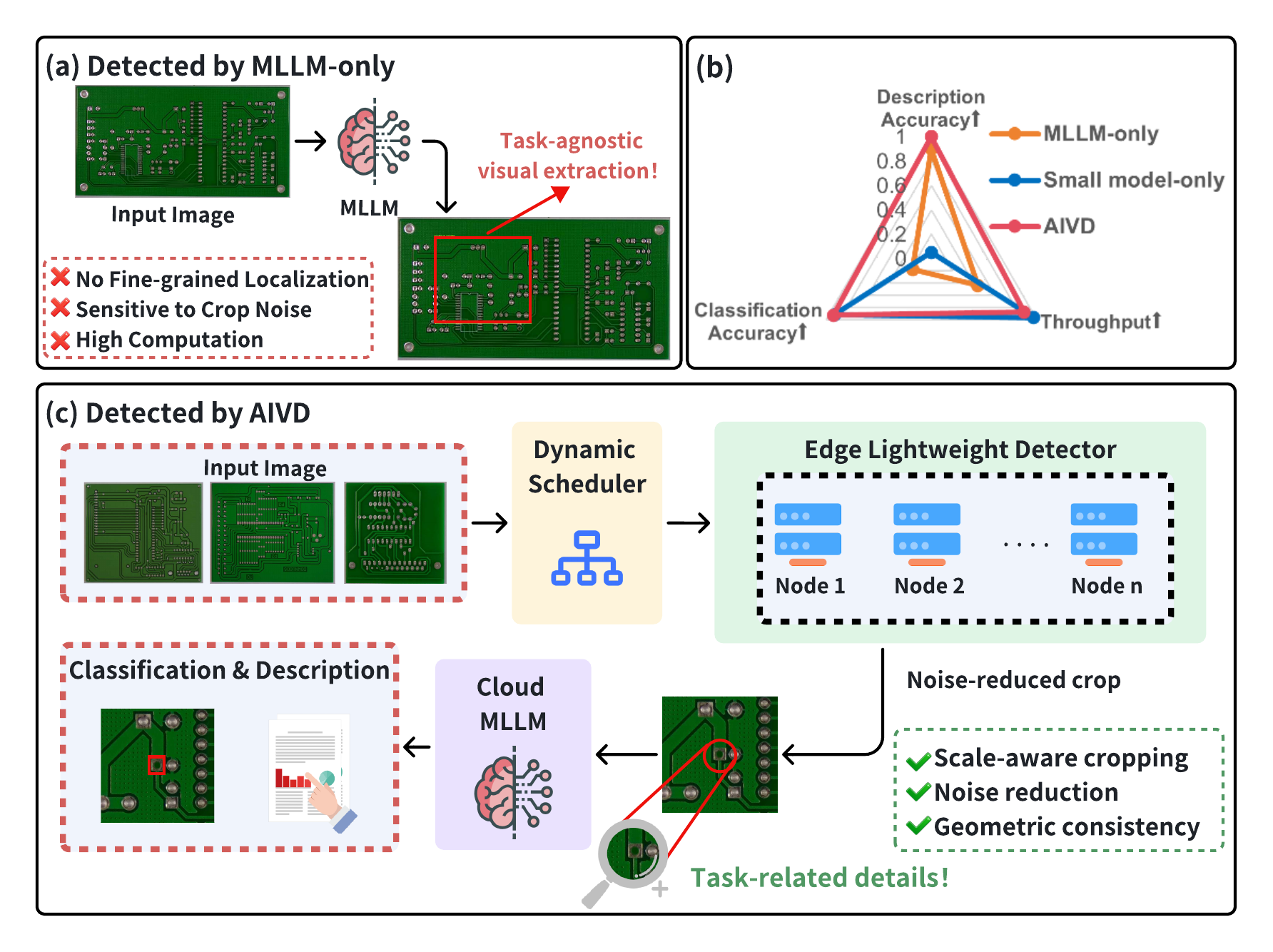}
    \caption{We propose AIVD, a novel framework for high-precision industrial visual detection. (a) Detected by MLLM-only: Struggle to meet precise localization demands in industrial scenarios. (b) Our approach outperforms existing methods. (c) Detected by AIVD: Achieves efficient object localization while generating precise defect cause descriptions.}
    \label{fig:fig1}
\end{figure}

In contrast, multimodal large language models (MLLMs) such as BLIP-2 \cite{3}, the LLaVA series~\cite{4,5}, and Qwen-VL \cite{6} possess robust visual-language joint reasoning capabilities. They can generate fine-grained descriptions and perform cross-modal reasoning, compensating for the semantic expression limitations of lightweight models. However, directly applying MLLMs to industrial-level visual detection still faces multiple challenges \cite{ADA-19}. Firstly, industrial defects are typically small-scale. It is difficult for MLLM to accurately identify these localized areas, particularly when impacted by background textures, process noise, and complex materials. Relying solely on the global perception of large models fails to meet the high-precision localization requirements for small targets, as shown in Fig. \ref{fig:fig1}(a). Secondly, MLLMs incur substantial computational overhead and long inference latency. This limitation makes them unsuitable for real-time deployment on edge devices and industrial production lines. The test results in Fig. \ref{fig:fig1}(b) demonstrate that lightweight models struggle to provide sufficient semantic expressiveness in large-scale industrial detection scenarios, while large models fail to meet real-time requirements. 

Recently, some studies have explored synergistic use of lightweight detectors and large models \cite{7}. DetGPT \cite{8} adopts a two-stage collaboration scheme to improve semantic understanding in open-scene settings. ContextDET \cite{9} employs large models to generate contextual information for the location of lightweight models. TaskCLIP \cite{10} uses semantic alignment to improve the relevance of tasks. Despite demonstrating the potential of large and small models' collaboration, these approaches exhibit limitations. Most of them focus primarily on semantic reasoning enhancement. They do not consider visual augmentation or domain adaptation for industrial scenarios involving small targets, heavy noise, and dense defects. In addition, their collaborative architectures often assume single-node deployment or static resource availability. As a result, they struggle to operate effectively in multi-edge environments with heterogeneous computing capabilities and dynamic network conditions.

To address these challenges, we propose the AIVD framework, as illustrated in Fig. \ref{fig:fig1}(c). It enhances MLLMs' accuracy in small-object scenarios through visual-semantic collaborative fine-tuning strategies, while achieving adaptive task allocation and stable inference performance in heterogeneous multi-edge environments via resource-aware scheduling. The main contributions of this paper are as follows:

\begin{itemize}

  \item  We propose AIVD, an adaptive edge-cloud collaborative framework for industrial visual detection. By the synergy of large and small models across the cloud and edge and heterogeneous resource-aware dynamic scheduling, AIVD can achieve accurate industrial visual detection and semantic reasoning under resource constraints.

  \item We design an efficient fine-tuning strategy for visual-semantic synergistic enhancement. It effectively aligns localized visual cues with semantic representations, significantly improving classification accuracy and semantic consistency.

  \item Extensive experiments across diverse industrial scenarios demonstrate that AIVD consistently outperforms baseline methods in both accuracy and efficiency. Compared to cloud-only solutions, throughput increased by an average of approximately 42.6\%, while resource consumption decreased by an average of approximately 13.5\%. Compared to traditional edge-cloud schemes, latency decreased by an average of approximately 15\%. 

\end{itemize}

\section{Proposed Method}
\subsection{Overall Design}

\begin{figure}[th]
    \centering
    \includegraphics[width=1\linewidth]{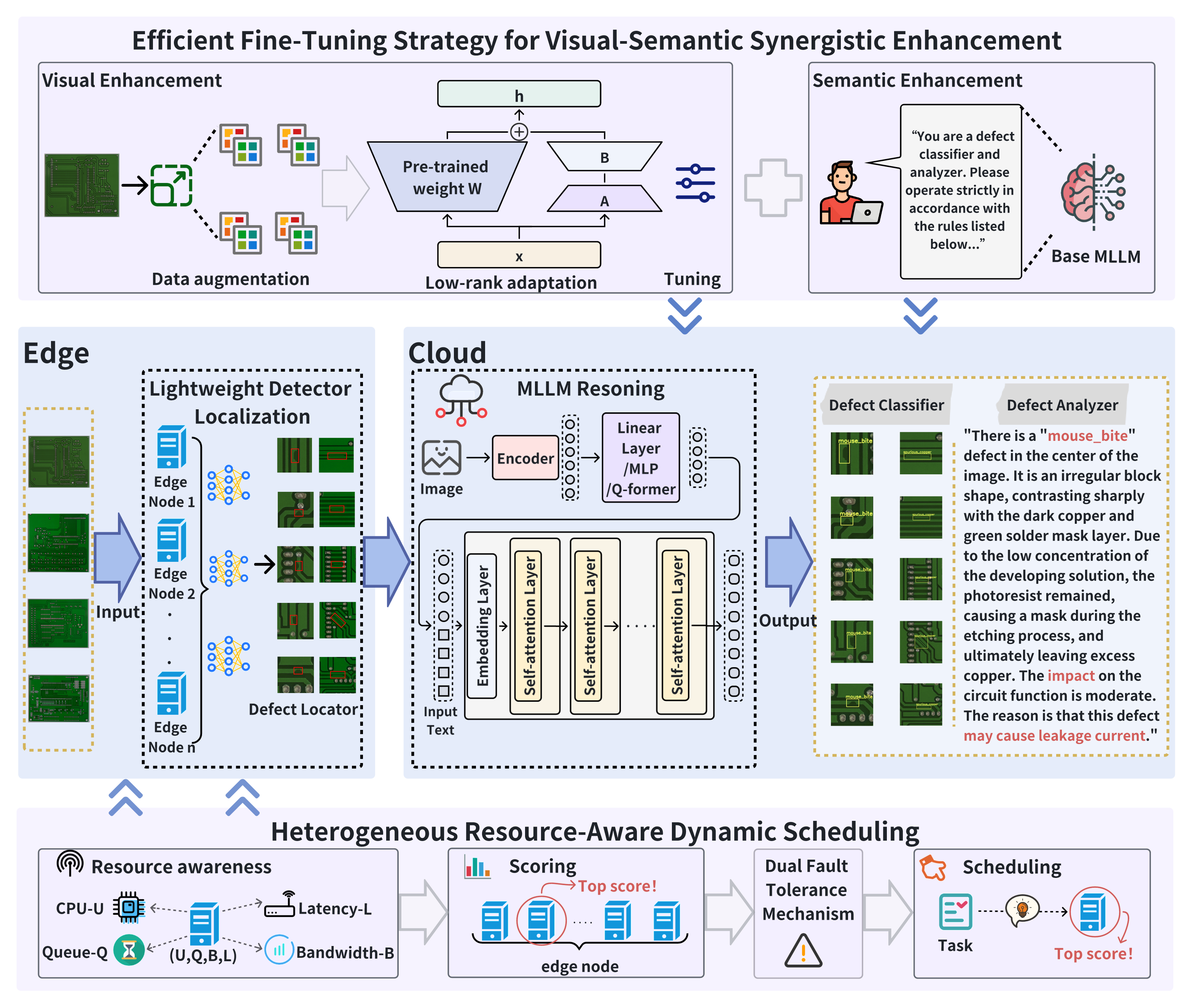}
    \caption{Overall framework of AIVD.}
    \label{fig:fig2}
\end{figure}

As shown in Fig. \ref{fig:fig2}, AIVD adopts edge-cloud collaboration as its core architecture. It jointly considers visual detection, semantic reasoning, and resource scheduling at the system level. This design aims to balance low latency with high semantic quality in large-scale industrial visual detection. The system comprises multiple heterogeneous edge nodes and a cloud-based multimodal inference node, with global coordination achieved through unified task management and monitoring services. Each edge node deploys a lightweight object detection model, responsible for performing high-frequency, low-cost local defect localization on input images. The cloud node deploys a multimodal large model, responsible for fine-grained classification of candidate local regions and structured semantic generation. Overall, the architecture can be viewed as a three-tier pipeline. The edge layer handles rapid localization through high-frequency visual detection. The cloud layer performs low-frequency inference with rich semantic reasoning. The system-level dynamic scheduling layer optimizes global performance under multi-node and multi-resource constraints.

\subsection{Efficient Fine-Tuning for Visual-Semantic Synergistic Enhancement}
Industrial defect images are typically characterized by dense local textures, contextual gaps, and small defect scales. These factors cause MLLM to suffer from semantic drift, category confusion, and weak causality inference capabilities. To address this, we propose an efficient fine-tuning strategy that synergistically enhances visual and semantic aspects. During fine-tuning, it simultaneously optimizes visual consistency, semantic discriminability, and cross-modal mapping capabilities, thereby significantly improving the model's robust recognition of diverse defect patterns.

In the original data, defect regions often occupy only a small proportion of the full image, preventing MLLM from establishing stable spatial-semantic correspondences during the visual encoding phase. We reconstruct visual inputs through dynamic context cropping and multidimensional image enhancement. Given an image $I$ and a defect bounding box $b_{i}$, we construct a cropping operator based on a context-expanded kernel:
\begin{equation}
x_{i}=\mathcal{C}\left(I, \mathcal{E}\left(b_{i} ; k\right)\right), \label{eq1}
\end{equation}
where $\mathcal{C}(\cdot)$ denotes the cropping operator, and $\mathcal{E}\left(b_{i} ; k\right)$ represents the context expansion kernel applied to bounding box $b_{i}$ with expansion rate $k$:
\begin{equation}
\mathcal{E}\left(b_{i} ; k\right)=b_{i} \oplus \mathcal{K}_{k}, \label{eq2}
\end{equation}
where $\oplus$ indicates boundary expansion to ensure strict geometric consistency of the expansion window. $\mathcal{K}_{k}$ is a structured expansion kernel scaled by $k$, which is used to balance the local details and global semantic information under different defect scales with the model's sensory field conditions. In this way, the model can establish a more stable spatial-semantic correspondence in the visual encoding stage and reduce the semantic uncertainty caused by the restricted field of view.

To enhance the robustness of the model against variations in lighting, texture, and scale in real industrial scenarios, we further introduce a set of combined data augmentation operators $\mathcal{A}(\cdot)$ and uniformly model brightness, contrast, saturation, and hue. Specifically, the augmented region $x_{i}$ is represented as $\tilde{x}_{i}=\mathcal{A}\left(x_{i} ; \theta_{a}\right)$. Where $\theta_{a}$ encompasses continuous perturbation parameters such as brightness, contrast, saturation, and hue. For the unified modeling of illumination and saturation, we separately sample scaling factors $\alpha$ and $\beta$ from a uniform distribution and define the diagonal scaling operator:

\begin{equation}
D(\alpha, \beta)=\operatorname{diag}(1, \alpha, \beta), \quad \alpha, \beta \sim \mathcal{U}(\cdot). \label{eq4}
\end{equation}

Additionally, define the projection operator $\Pi_{[0,255]}(\cdot)$ that restricts pixels to the valid range [0,255]. The enhanced HSV representation can be written as:
\begin{equation}
\widetilde{\mathrm{HSV}}_{i}=\Pi_{[0,255]}\left(D(\alpha, \beta) \operatorname{HSV}\left(x_{i}\right)\right). \label{eq5}
\end{equation}

This collaborative perturbation can simulate real-world industrial imaging factors such as lighting variations, camera exposure shifts, and saturation inconsistencies at low computational cost. This enables visual encoders to obtain more stable and generalizable visual representations during cross-modal alignment.

To further reduce semantic drift during inference in MLLMs, we introduce semantic prompt enhancement. During training, we construct diagnostic semantic descriptions for each sample: $t_{i}=\operatorname{Prompt}\left(c_{i}, d_{i}\right)$. Where $c_{i}$ denotes the defect category, and $d_{i}$ represents the manually designed description of the defective mechanism. Semantic enhancement enables the model to accomplish implicit semantic clustering during training by structurally describing the defect causes. That is, the same categories have consistent semantic contexts; the semantic distinction boundaries of similar categories are explicitly enhanced to construct robust visual-semantic mappings.


Based on visual and semantic enhancement, we use Low-Rank Adaptation (LoRA) to efficiently fine-tune the parameters of the MLLM. Given the linear transformation matrix in the original model: $h=Wx$. LoRA introduces low-rank increments to replace them in the update process: $h^{\prime}=(W+\Delta W) x, \Delta W=B A$. If $W \in \mathbb{R}^{d \times d}$, then LoRA decomposes it as:
\begin{equation}
A \in \mathbb{R}^{r \times d}, B \in \mathbb{R}^{d \times r}, r \ll d. \label{eq9}
\end{equation}



To further improve the expressiveness, we add a scaling factor $\lambda $ to the LoRA incremental parameters in the training phase to obtain the equation as follows:
\begin{equation}
W^{*}=W+\lambda \Delta W=W+\lambda B A. \label{eq10}
\end{equation}

To significantly reduce the number of new parameters added, only $A$ and $B$ are optimized during backpropagation. To control gradient explosion during the early training phase, we employ a zero-initialization strategy: $B=0, A \sim \mathcal{N}\left(0, \sigma^{2}\right)$. It can ensure $\Delta W=0$ and that the model's initial behavior matches the original model. Additionally, we introduce a regularization term $\mathcal{L}_{\text {LoRA-reg }}=\|A\|_{F}^{2}+\|B\|_{F}^{2}$ to prompt more stable and structured feature learning within the low-rank subspace.


\begin{table*}[t]
\centering
\caption{Accuracy comparison of different fine-tuning methods.}
\label{tab:tab1}
\footnotesize
\begin{tabular}{l l c c c c}
\hline
\textbf{MLLM} & \textbf{Method} & \textbf{DeepPCB Acc.} & \textbf{HriPCB Acc.} & \textbf{Average} & \textbf{Average Improvement} \\
\hline

\rowcolor{gray!15}
\multicolumn{6}{l}{\textbf{Zero-shot MLLMs}} \\
Qwen2-VL-7B & No fine-tuning & 0.167 & 0.218 & 0.193 & - \\
LLaVA1.6-mistral & No fine-tuning & 0 & 0 & 0 & - \\
InternVL3.5 & No fine-tuning & 0.171 & 0.295 & 0.233 & - \\

\hline
\rowcolor{gray!15}
\multicolumn{6}{l}{\textbf{Standard QLoRA\cite{16qlora}}} \\
Qwen2-VL-7B & QLoRA (8-bit), w/o Aug. & 0.545 & 0.524 & 0.535 & 0.342 \\
& QLoRA (8-bit), Rand-Aug (Rot./Sharp./ColorJit.) & 0.608 & 0.664 & 0.636 & 0.443 \\
LLaVA1.6-mistral & QLoRA (8-bit), w/o Aug. & 0.671 & 0.673 & 0.672 & 0.672 \\
& QLoRA (8-bit), Rand-Aug (Rot./Sharp./ColorJit.) & 0.791 & 0.719 & 0.755 & 0.755 \\
InternVL3.5 & QLoRA (8-bit), w/o Aug. & 0.913 & 0.935 & 0.924 & 0.691 \\
& QLoRA (8-bit), Rand-Aug (Rot./Sharp./ColorJit.) & 0.917 & 0.956 & 0.937 & 0.704 \\

\hline
\rowcolor{gray!15}
\multicolumn{6}{l}{\textbf{Standard LoRA\cite{17lora}}} \\
Qwen2-VL-7B & LoRA (r=8), w/o Aug. & 0.746 & 0.895 & 0.820 & 0.627 \\
& LoRA (r=8), Rand-Aug (Rot./Sharp./ColorJit.) & 0.777 & 0.908 & 0.843 & 0.650 \\
& LoRA (r=16), w/o Aug. & 0.704 & 0.907 & 0.806 & 0.613 \\
& LoRA (r=16), Rand-Aug (Rot./Sharp./ColorJit.) & 0.709 & 0.912 & 0.811 & 0.618 \\
LLaVA1.6-mistral & LoRA (r=8), w/o Aug. & 0.867 & 0.897 & 0.882 & 0.882 \\
& LoRA (r=8), Rand-Aug (Rot./Sharp./ColorJit.) & 0.870 & 0.887 & 0.879 & 0.879 \\
& LoRA (r=16), w/o Aug. & 0.843 & 0.891 & 0.867 & 0.867 \\
& LoRA (r=16), Rand-Aug (Rot./Sharp./ColorJit.) & 0.873 & 0.917 & 0.895 & 0.895 \\
InternVL3.5 & LoRA (r=8), w/o Aug. & 0.937 & 0.941 & 0.939 & 0.706 \\
& LoRA (r=8), Rand-Aug (Rot./Sharp./ColorJit.) & 0.948 & 0.952 & 0.949 & 0.716 \\
& LoRA (r=16), w/o Aug. & 0.946 & 0.949 & 0.948 & 0.715 \\
& LoRA (r=16), Rand-Aug (Rot./Sharp./ColorJit.) & 0.951 & 0.910 & 0.930 & 0.697 \\

\hline
\rowcolor{gray!15}
\multicolumn{6}{l}{\textbf{Ours}} \\
\textbf{Qwen2-VL-7B} & \textbf{Visual-Semantic Synergistic Enhancement} & \textbf{0.880} & \textbf{0.937} & \textbf{0.909} & \textbf{0.716} \\
\textbf{LLaVA1.6-mistral} & \textbf{Visual-Semantic Synergistic Enhancement} & \textbf{0.889} & \textbf{0.943} & \textbf{0.916} & \textbf{0.916} \\
\textbf{InternVL3.5} & \textbf{Visual-Semantic Synergistic Enhancement} & \textbf{0.968} & \textbf{0.975} & \textbf{0.972} & \textbf{0.739} \\

\hline
\end{tabular}
\end{table*}

\subsection{Heterogeneous Resource-Aware Dynamic Heuristic Scheduling}
Industrial edge-cloud collaborative environments exhibit significant heterogeneity and dynamism, with distinct nodes varying markedly in resource dimensions such as latency and bandwidth. Fixed scheduling strategies are highly prone to latency jitter or throughput degradation under high-load conditions. To address this, we propose a heterogeneous resource-aware dynamic scheduling algorithm that enables adaptive allocation of inference tasks across edge nodes, as illustrated in Algorithm \ref{alg:resource-aware-scheduling}.

\begin{algorithm}[t]
\caption{Heterogeneous Resource-Aware Dynamic Scheduling}
\label{alg:resource-aware-scheduling}
\begin{algorithmic}[1]
\REQUIRE Edge nodes $\{1, \ldots, i, \ldots, N\}$, task stream $T$, weights $\alpha, \beta, \delta, \varepsilon$, smoothing factor $\eta$
\ENSURE Dynamic assignment decisions and updated scores $S_i$
\STATE initialize $S_i \gets 0$ for all $i$
\FOR{each incoming task $t \in T$}
    \FOR{each node $i$}
        \STATE $R_i \gets \text{Monitor}(i) = \left[U_{i}, Q_{i}, B_{i}, L_{i}\right]^{T}$
        \STATE $\text{score}_i \gets \alpha U_i + \beta Q_i + \delta B_i + \varepsilon L_i$
        \STATE $S_i \gets \eta S_i + (1 - \eta) \cdot \text{score}_i$
        \IF{\text{OverloadDetected}$(R_i)$}
            \STATE $S_i \gets \gamma \cdot S_i$
        \ENDIF
        \STATE $S_i \gets \max(S_i, \epsilon)$
    \ENDFOR
    \STATE Assign task $t$ to node $i^* \gets \arg\max_{i \in N} S_i$
\ENDFOR
\end{algorithmic}
\end{algorithm}

For each edge node $i$, the system periodically collects four key operational metrics, which are normalized into the $[0,1]$ range to form the node's heterogeneous resources state vector ${R}_{i}=[U_{i}, Q_{i}, B_{i}, L_{i}]^\mathrm{T}$. Where $U_{i}$ denotes the CPU idleness reflecting the available computational headroom of node $i$, $Q_{i}$ represents the queue congestion score, $B_{i}$ measures the available bandwidth capacity, and $L_{i}$ captures the network latency. This unified representation enables consistent comparison across heterogeneous nodes with different physical capabilities. Based on the resource state vector $R_{i}$, we construct an adjustable linear fusion scoring function:
\begin{equation}
S_{i}=w^{\mathrm{T}} R_{i}=\alpha U_{i}+\beta Q_{i}+\delta B_{i}+\varepsilon L_{i}, \label{eq14}
\end{equation}
where $w^{\mathrm{T}}=[\alpha, \beta, \delta, \varepsilon]^{\mathrm{T}}$ is a scheduling hyperparameter used to regulate the weights and realize the balance of multi-objective performance.
To improve the timing stability of the algorithm, we introduce an exponential smoothing strategy to dynamically update the node scores based on the latest monitoring status:
\begin{equation}
S_{i}(t+1)=\eta S_{i}(t)+(1-\eta) \cdot f\left(R_{i}(t)\right), \label{eq15}
\end{equation}
where $\eta \in[0,1]$ controls the fusion ratio of old and new information, and $f\left(R_{i}(t)\right)$ denotes the instantaneous scores generated from the real-time monitoring metrics.

In the task allocation phase, the system adopts a deterministic greedy multi-indicator fusion selection mechanism, where each incoming task is assigned to the node $i^{*}$ with the highest aggregated scheduling score $i^{*}=\arg \max_{i \in N} S_{i}$, and $N$ denotes the set of available nodes. This strategy prioritizes the node that best matches the current global resource state at each scheduling step, thereby ensuring efficient convergence toward an optimal resource utilization regime while incurring minimal decision overhead.

To ensure the stable operation of the scheduling system under extreme loads and abnormal network conditions, we design a double fault tolerance mechanism. When monitoring a node's resource overload or delay abnormality, the system will automatically perform adaptive weight reduction processing on its score, so that it will be weakened rather than suddenly deleted in the scheduling:
\begin{equation}
S_{i}^{\text {penalized }}=\gamma(t)\cdot S_{i}(t), \gamma \in(0,1), \label{eq17}
\end{equation}
where $\gamma(t)$ is an adaptive attenuation factor that can be dynamically adjusted according to the severity of the detected anomaly.

In addition, to avoid the degradation of scoring to zero or negative values leading to unstable decision making, we introduce scoring lower bound protection:
\begin{equation}
S_{i}^{\prime}=\max \left(S_{i}^{\text {penalized }}, \epsilon\right), \epsilon=10^{-6}. \label{eq18}
\end{equation}

This strategy ensures that nodes retain minimum visibility even in extreme cases, thus achieving smooth and continuous scheduling dynamics.

\section{Experiments}
\subsection{Experimental Setup}
\textbf{Dataset.} In order to verify the effectiveness and system adaptability of AIVD in the specific application scenario of industrial defect detection, we selected two representative PCB industrial defect datasets: DeepPCB\cite{11deeppcb} and HRIPCB\cite{12hripcb}. 

\textbf{Implementation Details.} To evaluate the performance of AIVD in real edge-cloud heterogeneous environments, we build a distributed collaborative detection-reasoning system. The system consists of multiple heterogeneous edge nodes and one cloud server. The number of edge nodes is set to n = 4, 8, 12, 16. The edge nodes differ in computing capability and memory capacity. The memory limits range from 1 to 8 GB. Network conditions vary from 5 to 60 ms latency, with bandwidth between 20 and 200 Mbit/s. These configurations form a heterogeneous edge computing environment with highly imbalanced resources.
Each edge node runs YOLOv12s\cite{15yolov12} for detection, and the cloud node runs MLLMs such as Qwen2-VL-7B\cite{13qwen2}, LLaVA-V1.6-mistral-7B\cite{4}, and InternVL3.5\cite{14internvl} for semantic reasoning.
We further design three representative evaluation scenarios. Scenario 1 adopts normal network latency with low queue pressure. Scenario 2 introduces approximately 100 ms latency with moderate queue pressure. This setting simulates common network fluctuations and mild resource contention. Scenario 3 imposes 500 ms latency with high queue pressure. It is used to evaluate system performance under extremely constrained networks and heavy workloads.

\textbf{Evaluation Metrics.} AIVD is a system-level collaborative framework designed for industrial scenarios. Therefore, we evaluate it from two complementary perspectives: model performance and system performance. For model performance, we adopt classification accuracy as the primary metric. This metric reflects the ability of MLLMs to discriminate fine-grained defect categories. For system performance, we focus on key operational indicators relevant to industrial deployment. These indicators include resource consumption, end-to-end latency, and overall system throughput.

\begin{figure}[th]
    \centering
    \includegraphics[width=1\linewidth]{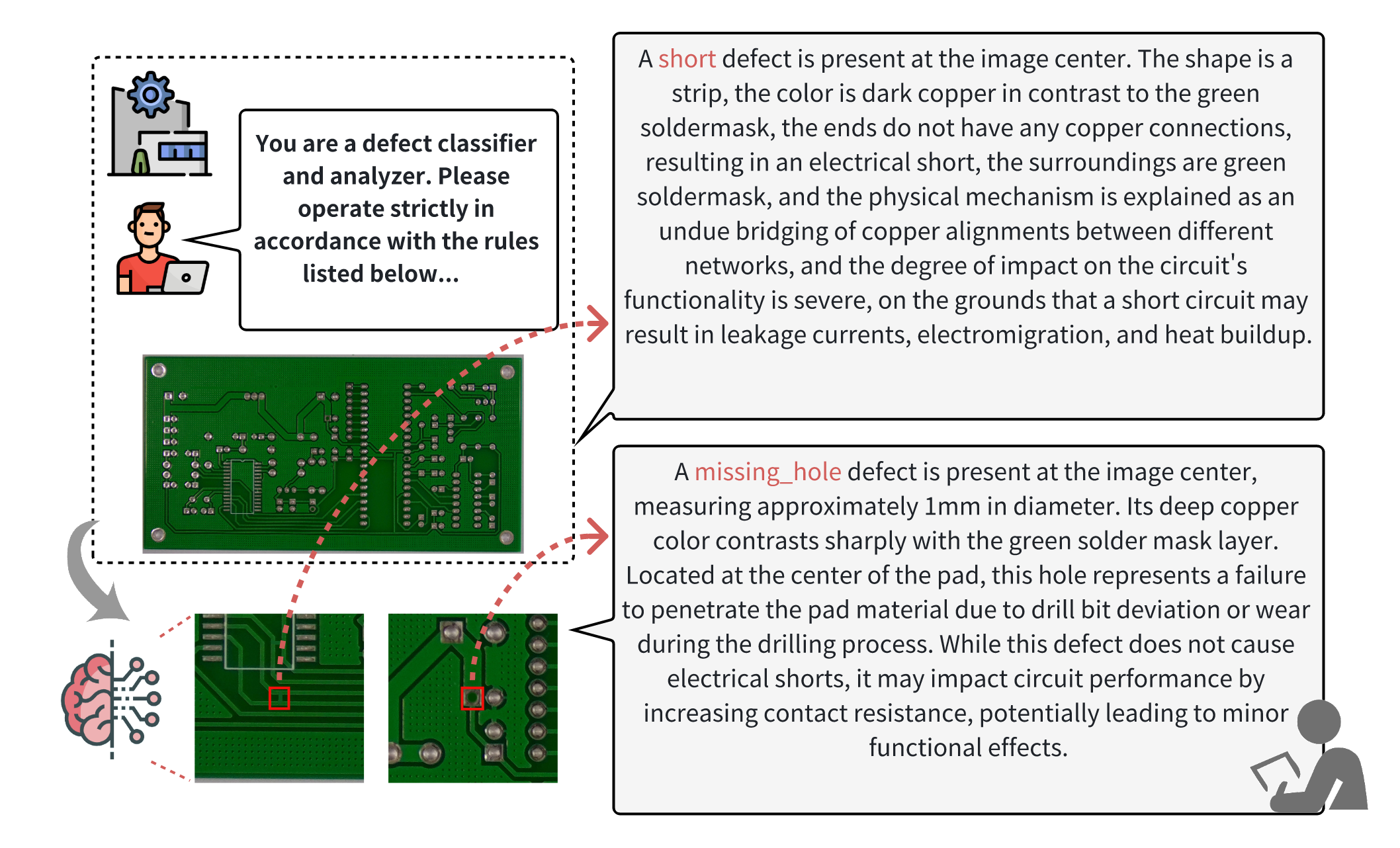}
    \caption{Example of AVID semantic analysis.}
    \label{fig:fig3}
\end{figure}

\subsection{Experimental Results}

\subsubsection{Fine-Tuning Performance of MLLM}
This section evaluates the effectiveness of different fine-tuning strategies in improving the visual understanding capability of MLLMs for industrial defect detection. Zero-shot MLLMs refer to base models without industrial prior knowledge. The setting “w/o Aug.” denotes tuning only through low-rank parameter updates, while “Rand-Aug (Rot. / Sharp. / ColorJit.)” introduces generic random augmentations to simulate image variations. The experimental results in Table \ref{tab:tab1} show that parameter fine-tuning and stochastic augmentation can improve the discriminative performance of MLLMs. However, these gains are inconsistent across different models and datasets. Moreover, such strategies struggle to reliably resolve semantic confusion among fine-grained defect categories. 
In comparison, the proposed visual-semantic synergistic enhancement and fine-tuning strategy achieves optimal classification accuracy on both datasets and multiple MLLM architectures. Fig. \ref{fig:fig3} demonstrates the semantic output quality of AIVD. With the proposed fine-tuning strategy, the MLLM accurately captures fine-grained defect morphology while preserving strong semantic consistency. 
In addition, the model produces coherent and interpretable explanations of potential defect causes. This behavior indicates improved visual-semantic alignment. These results confirm that the fine-tuning strategy enhances semantic robustness under imperfect localization, enabling reliable industrial defect interpretation.


\subsubsection{System Performance}
\begin{figure}[h]
	
	\begin{minipage}{0.32\linewidth}
		\vspace{3pt}
		\centerline{\includegraphics[width=\textwidth]{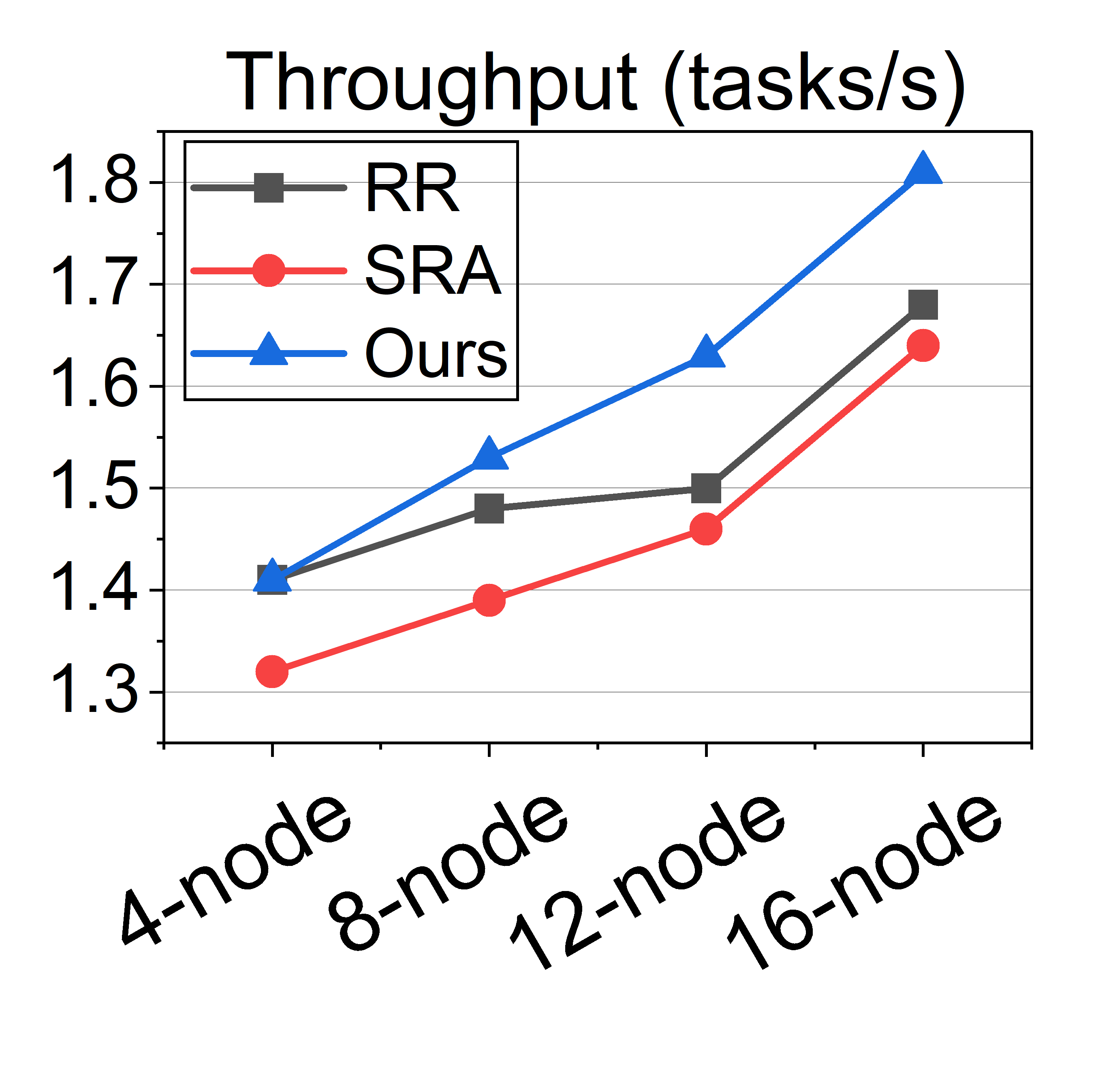}}
		\centerline{(a) Scenario 1}
            \label{fig:fig4a}
	\end{minipage}
	\begin{minipage}{0.32\linewidth}
		\vspace{3pt}
		\centerline{\includegraphics[width=\textwidth]{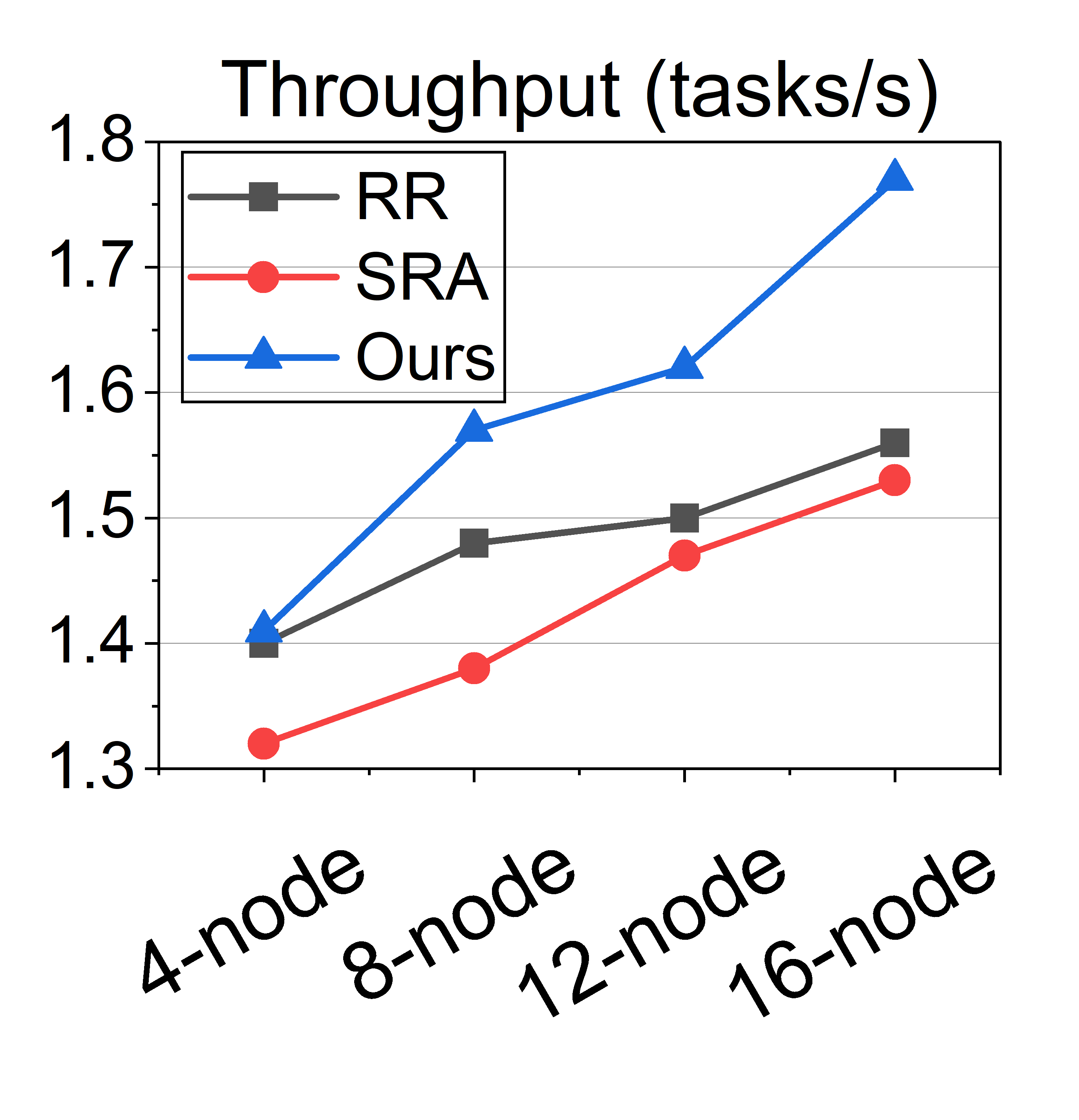}}
		\centerline{(b) Scenario 2}
            \label{fig:fig4b}
	\end{minipage}
	\begin{minipage}{0.32\linewidth}
		\vspace{3pt}
		\centerline{\includegraphics[width=\textwidth]{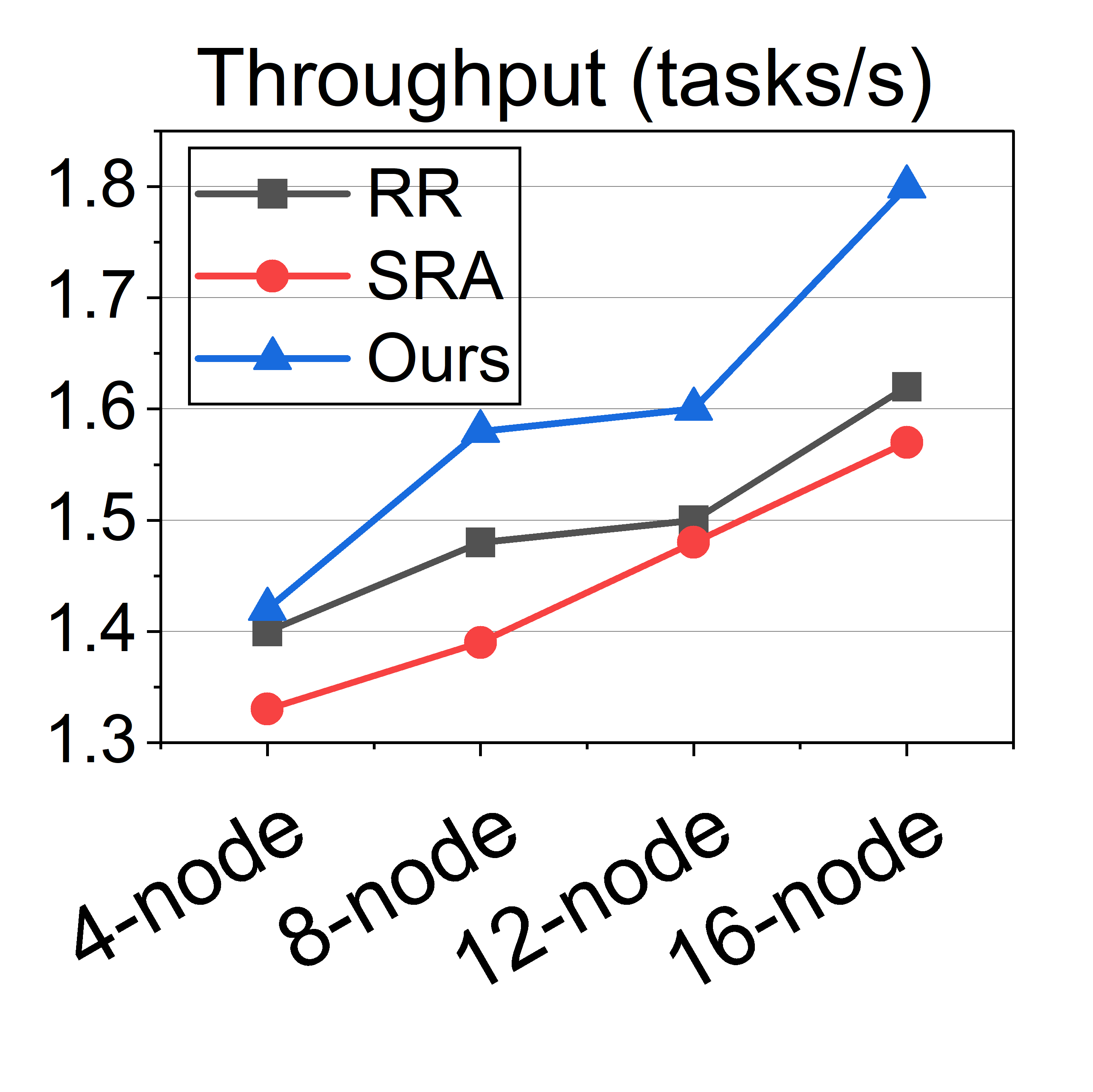}}
		\centerline{(c) Scenario 3}
            \label{fig:fig4c}
	\end{minipage}
 
    \caption{Comparison of edge-side throughput under different numbers of nodes.}
	\label{fig:fig4}
\end{figure}

\begin{figure}[h]
	
	\begin{minipage}{0.32\linewidth}
		\vspace{3pt}
		\centerline{\includegraphics[width=\textwidth]{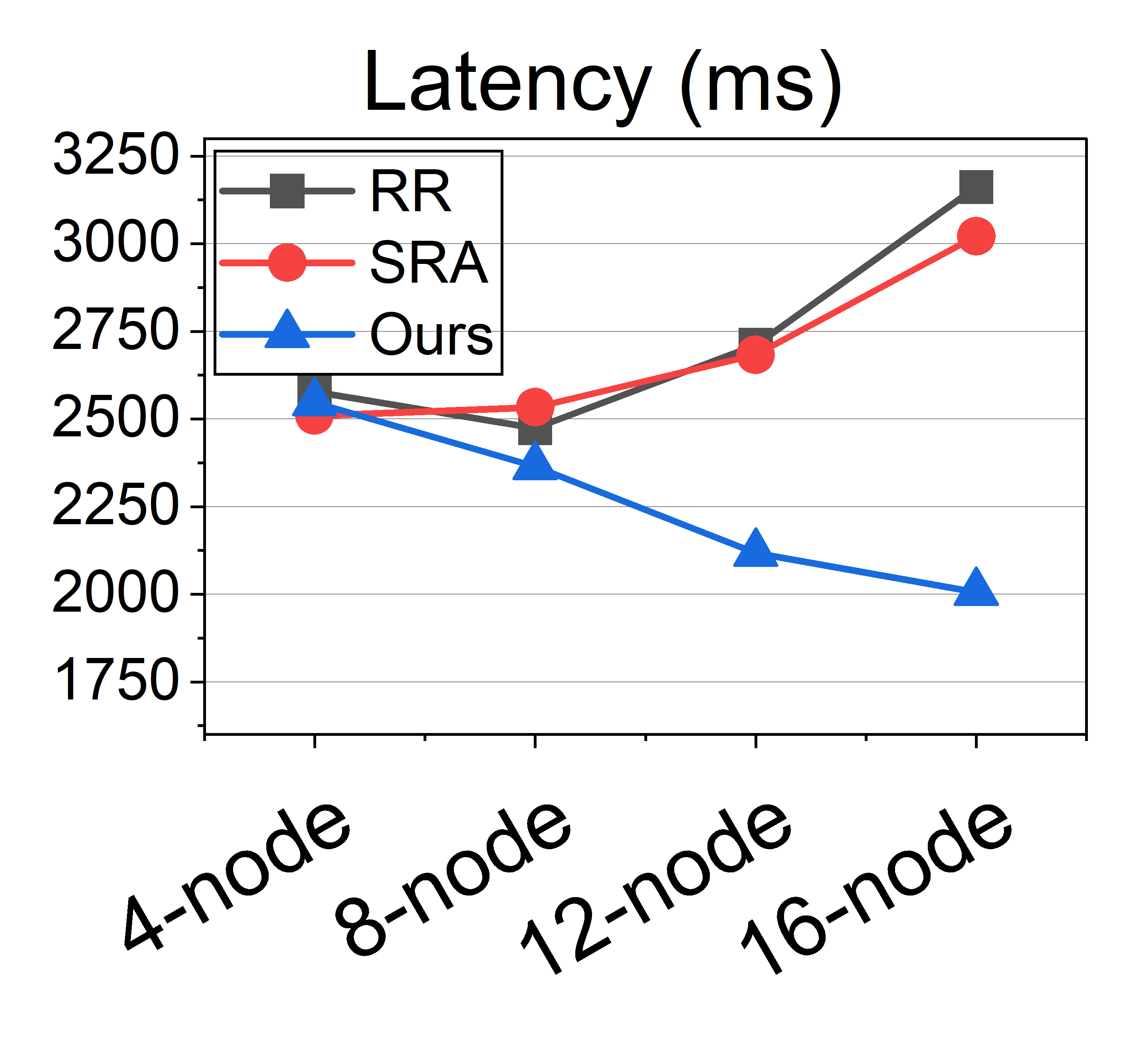}}
		\centerline{(a) Scenario 1}
            \label{fig:fig5a}
	\end{minipage}
	\begin{minipage}{0.32\linewidth}
		\vspace{3pt}
		\centerline{\includegraphics[width=\textwidth]{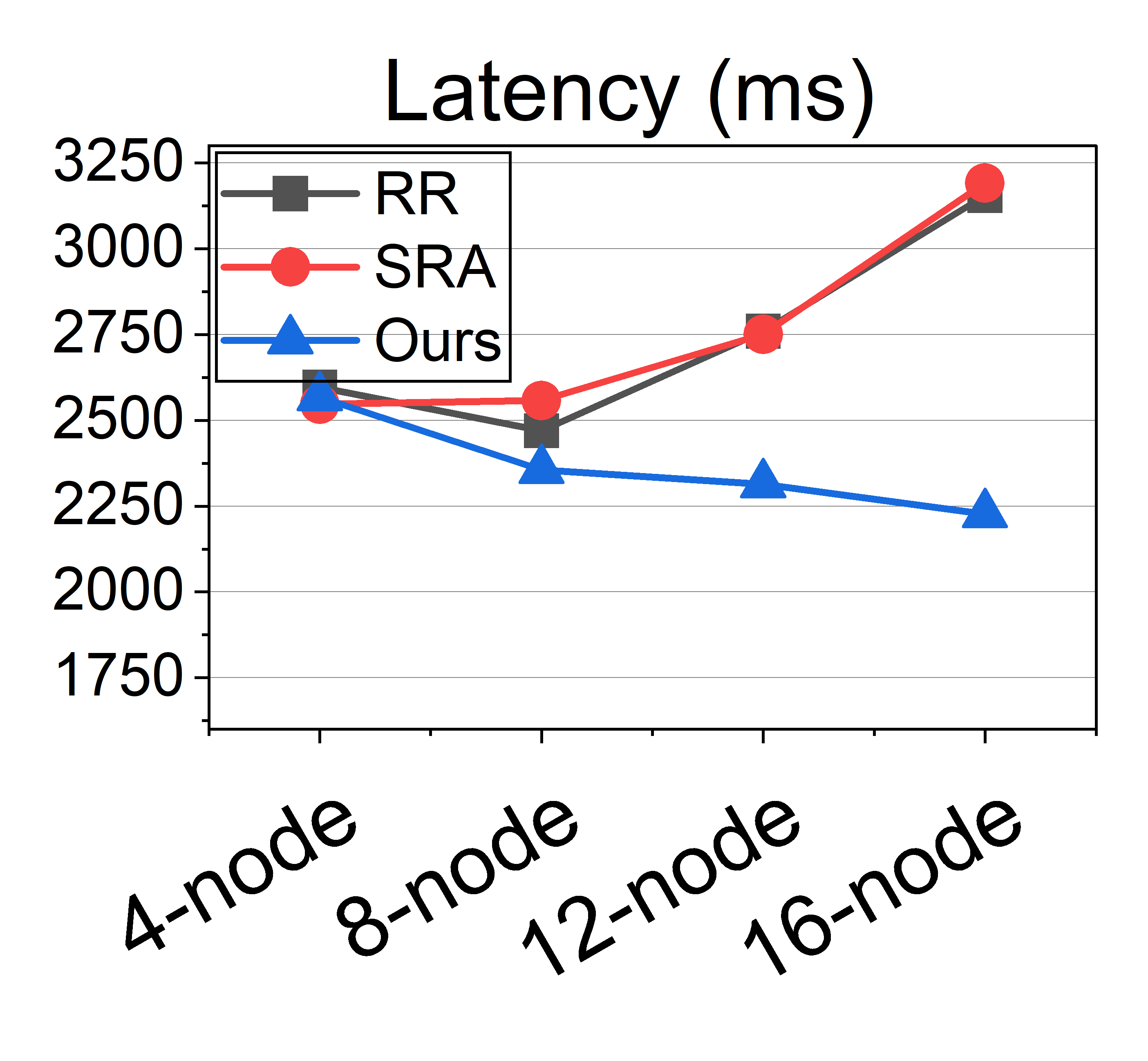}}
		\centerline{(b) Scenario 2}
            \label{fig:fig5b}
	\end{minipage}
	\begin{minipage}{0.32\linewidth}
		\vspace{3pt}
		\centerline{\includegraphics[width=\textwidth]{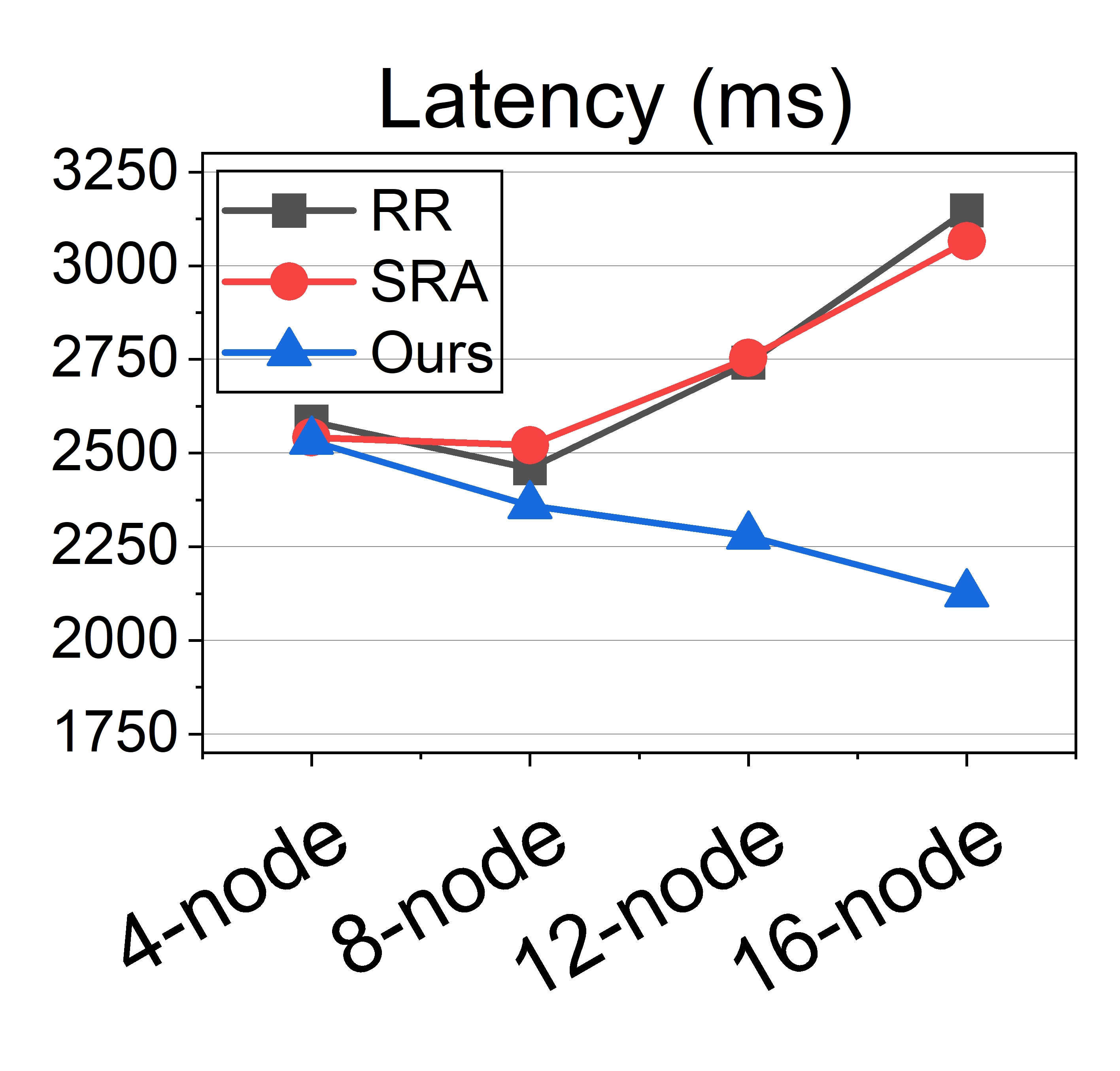}}
		\centerline{(c) Scenario 3}
            \label{fig:fig5c}
	\end{minipage}
 
    \caption{Comparison of edge-side latency under different numbers of nodes.}
	\label{fig:fig5}
\end{figure}

Fig. \ref{fig:fig4} and \ref{fig:fig5} present the throughput and average latency of different scheduling strategies under varying node scales. We compare the proposed scheduling strategy with the edge-side Round-Robin strategy (RR) and the static resource-aware allocation policy (SRA) based on CPU idleness. With more nodes, heterogeneity exacerbates bottlenecks. For example, under network constraints, RR and SRA methods overload low‑performance nodes, spiking the queue backlog and tail latency. In contrast, our strategy consistently outperforms both baselines. Especially in Scenario 3, the throughput with 16 nodes improves by 11.1\% and 14.6\% compared to RR and SRA, respectively. The throughput advantage becomes more evident as the node scale increases. This trend indicates that the proposed strategy can effectively exploit high-performance nodes in heterogeneous environments. Moreover, the reduction in average latency indicates improved load-balancing stability rather than short-term burst optimization. By proactively assigning tasks to nodes with sufficient processing headroom, the strategy reduces queue oscillation and mitigates tail latency under fluctuating network conditions. 



\begin{figure}[h]
	\centering
	\begin{minipage}{0.48\linewidth}
		\vspace{3pt}
		\centerline{\includegraphics[width=\textwidth]{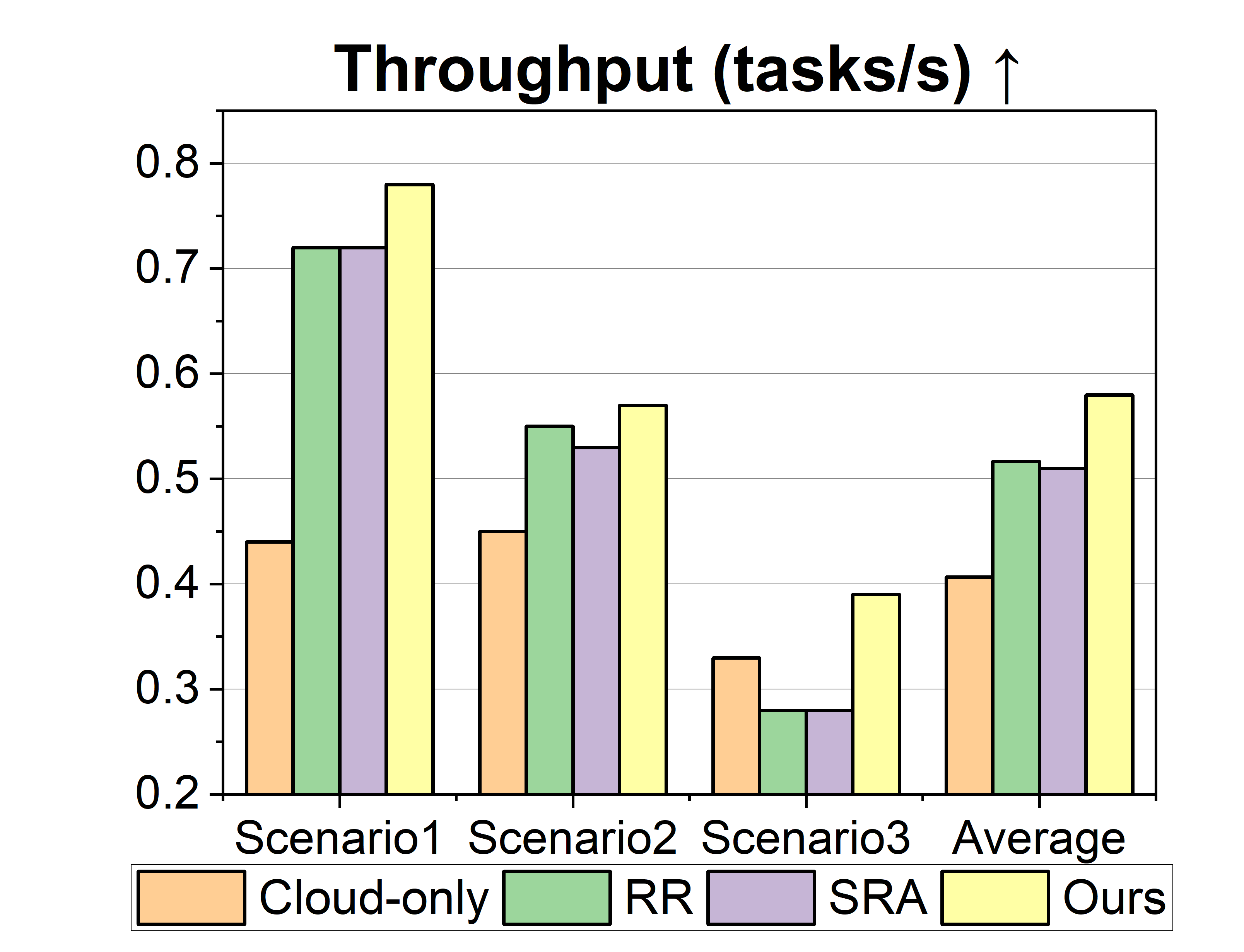}}
		\centerline{(a) Throughput}
		\label{fig:fig6a}
	\end{minipage}
	\hfill
	\begin{minipage}{0.48\linewidth}
		\vspace{3pt}
		\centerline{\includegraphics[width=\textwidth]{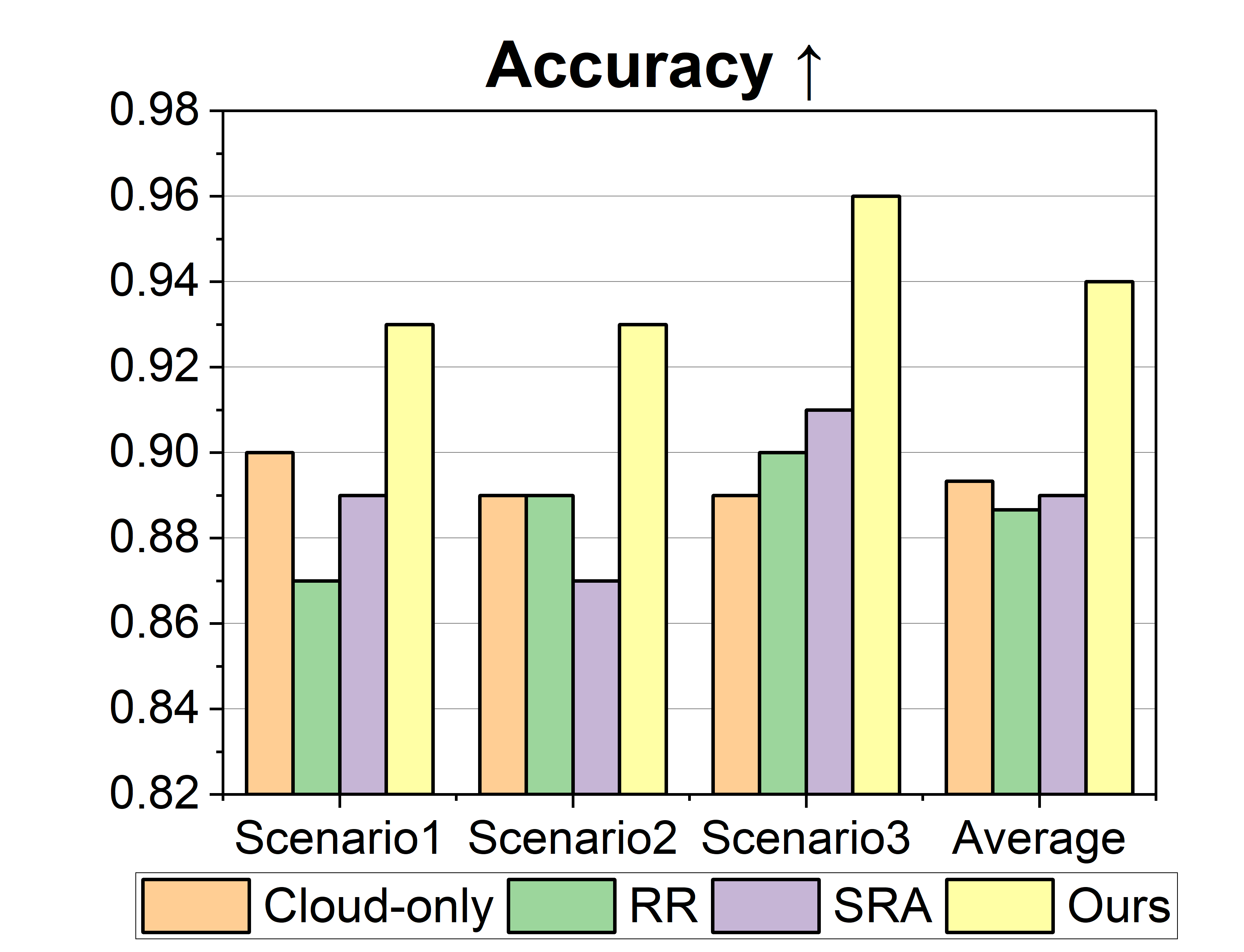}}
		\centerline{(b) Accuracy}
		\label{fig:fig6b}
	\end{minipage}

	\vspace{6pt}

	\begin{minipage}{0.48\linewidth}
		\vspace{3pt}
		\centerline{\includegraphics[width=\textwidth]{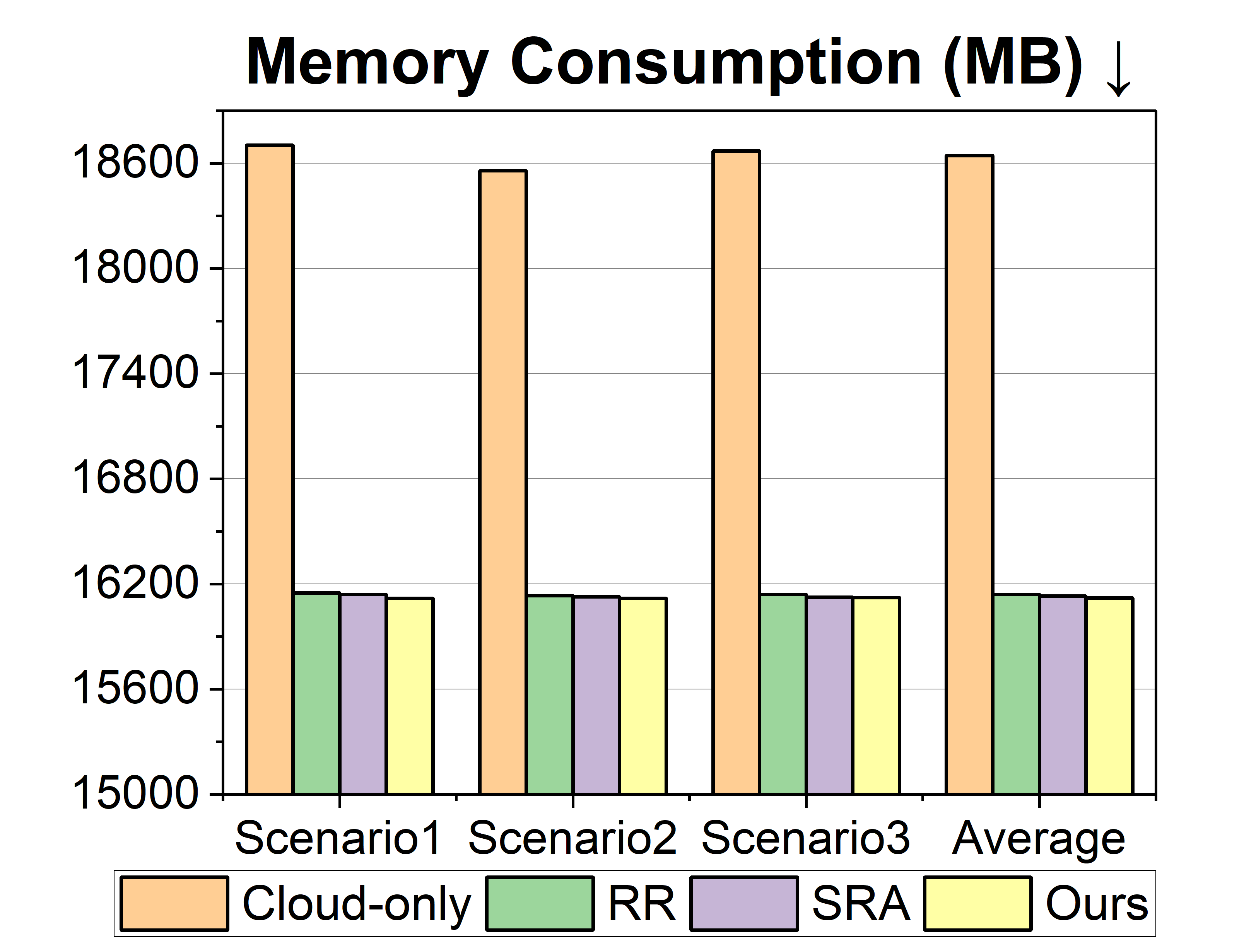}}
		\centerline{(c) Memory Consumption}
		\label{fig:fig6c}
	\end{minipage}
	\hfill
	\begin{minipage}{0.48\linewidth}
		\vspace{3pt}
		\centerline{\includegraphics[width=\textwidth]{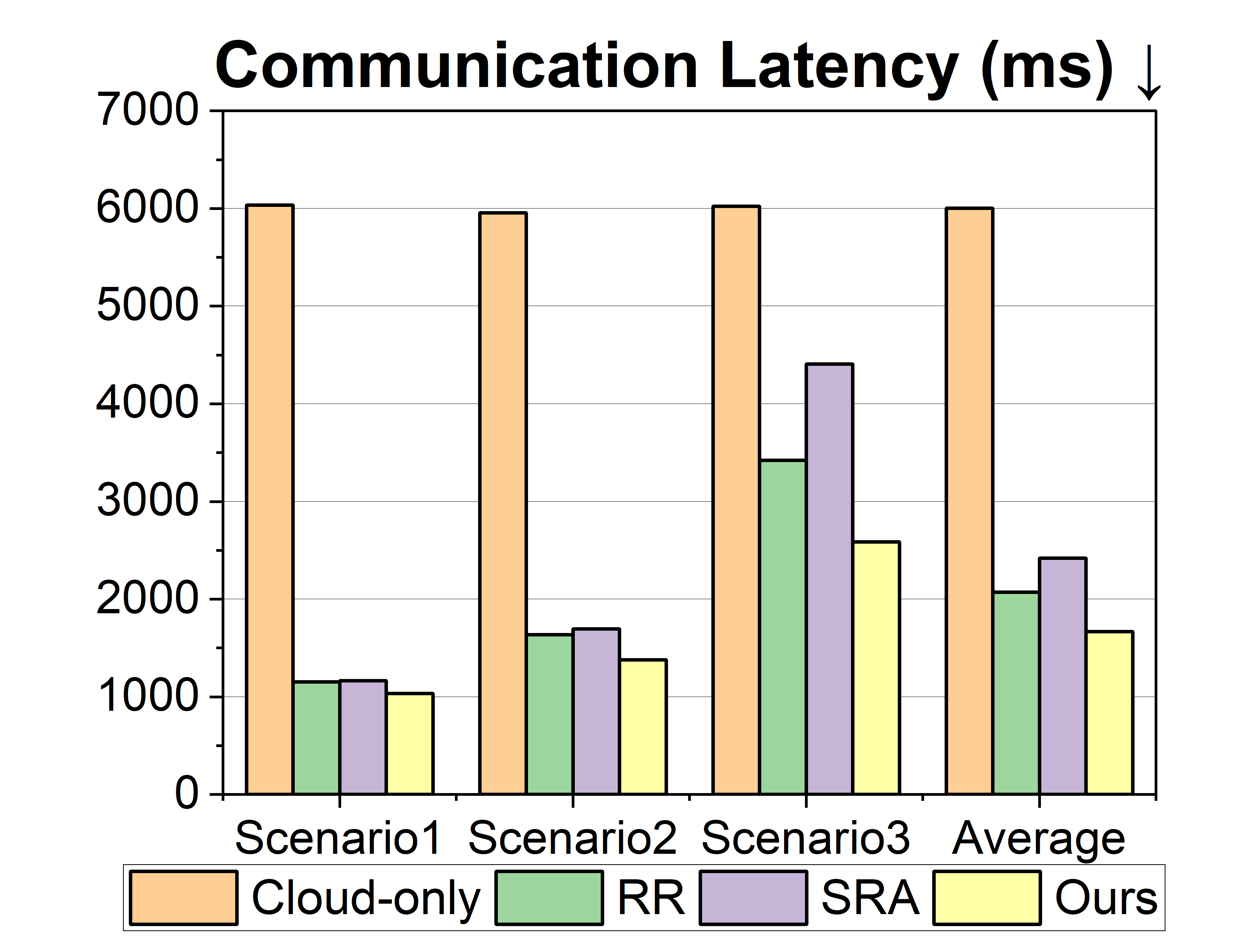}}
		\centerline{(d) Communication Latency}
		\label{fig:fig6d}
	\end{minipage}

	\caption{The end-to-end system performance comparison results. ↑ represents higher is better, ↓ represents lower is better.}
	\label{fig:fig6}
\end{figure}

In addition, our strategy can stabilize the average latency in multiple scenarios by jointly considering the arithmetic occupation, network latency, and cache pressure. Under scenarios 1 and 3 with 16 nodes, the average latency of our proposed strategy is reduced by 36.5\% and 32.5\% compared to RR and SRA, respectively. 
Fig. \ref{fig:fig6} shows that the complete AIVD framework exhibits optimal overall performance and stability in all scenarios. Specifically, in scenario 1, AIVD improves the throughput by 77\% compared to the Cloud-only scheme, increases the accuracy to 0.93 and reduces the average resource consumption by about 13.8\%. Under the network fluctuation conditions in scenarios 2 and 3, AIVD still maintains the highest throughput rate with classification accuracy while significantly reducing the memory. In addition, AIVD reduces the communication delay by 57.1\%, 24.4\%, and 41.3\% compared to the Cloud-only, RR, and SRA in scenario 3. 
These results indicate that the performance gains of AIVD are not limited to a single metric but arise from a balanced optimization across computation, communication, and memory resources. The proposed strategy avoids local overload and excessive offloading, leading to more stable latency behavior under varying workloads. This explains the consistent reduction in average delay across multiple scenarios, particularly in large-scale deployments with 16 nodes. 



\section{Conclusion}
In this paper, we propose AIVD, which is an adaptive large-small model synergy framework for edge-cloud collaboration in industrial visual detection. The framework targets high-precision localization, robust semantic understanding, and low-latency inference under resource constraints. The framework effectively mitigates the challenges of small scale, strong background noise and semantic drift in industrial defect images by working with lightweight detectors and MLLMs, combined with an efficient fine-tuning strategy of visual-semantic synergistic enhancement. The introduced heterogeneous resources-aware dynamic scheduling algorithm enables the system to improve throughput and reduce latency under heterogeneous arithmetic and fluctuating network conditions. Experimental results demonstrate that AIVD can reduce resource consumption while improving MLLM classification performance and semantic generation quality. 


\bibliographystyle{IEEEbib}
\bibliography{icme2026references}


\end{document}